\pdfoutput=1
\documentclass[sigconf,nonacm]{acmart}
\usepackage{mathtools}
\usepackage{booktabs}
\usepackage{tikz}
\usepackage{multirow}
\usepackage{graphicx}
\usepackage{amsmath}
\usepackage{amsthm}

\setcopyright{none}
\acmDOI{}
\acmISBN{}



\title[Biquaternionic Space with Complex-valued Attention for TKGC]{Biquaternionic Space with Complex-valued Attention for Temporal Knowledge Graph Completion}

\author{Rushan Geng}
\affiliation{%
  \department{School of Computer Science and Technology}
  \institution{University of Chinese Academy of Sciences}
  \city{Beijing}
  \country{China}
}
\email{gengrushan23@mails.ucas.ac.cn}

\author{Cuicui Luo}
\affiliation{%
  \department{International College}
  \institution{University of Chinese Academy of Sciences}
  \city{Beijing}
  \country{China}
}
\email{luocuicui@ucas.ac.cn}
\renewcommand{\shortauthors}{Geng and Luo}

\begin{document}

\begin{abstract}
Temporal knowledge graph embedding (TKGE) models infer missing facts in knowledge graphs that evolve over time. Many existing models use a single geometric space, which can limit their ability to represent diverse relational patterns, or treat entity representations as static. We propose Biquaternionic Space with Complex-valued Attention (BSCA), a TKGE model that combines circular and hyperbolic rotations within a unified biquaternionic framework. A complex-valued attention mechanism adaptively fuses time-conditioned and relation-conditioned entity representations, allowing them to vary with temporal and relational context. Experiments on five benchmark datasets show competitive performance across datasets, with the largest improvement on GDELT: BSCA achieves an MRR of 52.1\%, compared with 38.1\% for the strongest baseline in our comparison.
\end{abstract}

\maketitle
\pagestyle{plain}

\section{Introduction}
Knowledge graphs (KGs) represent relations between real-world entities as triples $(s, r, o)$ and support applications such as question answering~\cite{saxena2020improving}, relation extraction~\cite{hu-etal-2021-knowledge}, and recommendation~\cite{he2023exploring}. Because facts can change over time, temporal knowledge graphs (TKGs) extend this representation to quadruples $(s, r, o, \tau)$, where $\tau$ indicates when a fact holds. For example, (Barack Obama, Make a visit, Malaysia, 2014-04-25) associates a visit with a specific date. TKGs are typically incomplete, motivating temporal knowledge graph embedding (TKGE) methods that infer missing facts from observed temporal information.

Existing TKGE approaches, including Euclidean models~\cite{leblay2018deriving,dasgupta2018hyte} and models based on complex or quaternion representations~\cite{lacroix2019tensor,chen2022rotateqvs}, achieve strong completion performance. Two modeling challenges nevertheless remain. First, different relational patterns favor different geometric transformations. Hyperbolic representations are well suited to hierarchical structure~\cite{Balazevic2019Poincare,chami2020low}, whereas circular rotations provide a natural representation of periodic patterns. Combining these transformations within a unified model can therefore be useful. Second, approaches that incorporate time only into relations leave entity representations static, limiting their ability to capture changes in entity semantics.

We address these challenges with Biquaternionic Space with Complex-valued Attention (BSCA). BSCA uses the Hamilton product of biquaternions to combine circular rotations, hyperbolic rotations, and scaling within a unified algebraic framework. This formulation provides flexible geometric transformations for different relations. A complex-valued attention mechanism then fuses time-conditioned and relation-conditioned entity representations, allowing their contributions to depend on the temporal and relational context. We analyze the model's ability to represent key relation patterns and evaluate its performance on five benchmarks.

Our main contributions are as follows:
\begin{itemize}
	\item We propose BSCA, a TKGE model in biquaternionic space that uses the non-commutative Hamilton product to simultaneously model circular and hyperbolic rotations within a unified algebraic framework.
	\item We introduce a complex-valued attention mechanism that dynamically fuses time-conditioned and relation-conditioned entity representations, enabling entities to evolve over time and across relations.
	\item We evaluate BSCA on five benchmarks. On GDELT, BSCA improves MRR by approximately 37\% relative to the strongest baseline included in our comparison.
\end{itemize}

\section{Related Work}
\subsection{Static Knowledge Graph Completion}
TransE~\cite{bordes2013translating} models a relation as a translation between entity embeddings, using the distance $\|e_s + e_r - e_o\|_p$, where $e_s$ and $e_o$ are entity vectors and $e_r$ is a relation vector. TransH~\cite{wang2014knowledge}, TransR~\cite{lin2015learning}, and TransD~\cite{ji2015knowledge} extend this approach through projections onto relation-specific hyperplanes or spaces. ComplEx~\cite{trouillon2016complex} instead uses tensor factorization to score facts in complex space. RotatE~\cite{sun2018rotate}, QuatE~\cite{zhang2019quaternion}, and DualE~\cite{cao2021dual} represent relations through transformations in complex, quaternion, and dual quaternion spaces, respectively. Neural models such as ConvE~\cite{dettmers2018convolutional} and InteractE~\cite{vashishth2020interacte} learn interactions between embeddings without explicitly prescribing a geometric transformation. BiQUE~\cite{guo2021bique} introduces biquaternionic representations for static KGE, while DGS~\cite{Iyer2022DGS} models interactions between two geometric spaces. These approaches motivate the use of multiple geometric transformations in knowledge graph representations.

\subsection{Temporal Knowledge Graph Completion}
TKGE models incorporate time into representations and scoring functions for temporal facts. TTransE~\cite{leblay2018deriving} extends TransE by representing timestamps as translations. HyTE~\cite{dasgupta2018hyte} associates temporal information with hyperplanes, and ATiSE~\cite{xu2020temporal} models trend, seasonal, and random components using Gaussian distributions. TComplEx~\cite{lacroix2019tensor} extends complex-valued tensor factorization to a fourth-order temporal tensor. TeRo~\cite{xu2020tero} represents temporal changes in entities as rotations in complex space, whereas ChronoR~\cite{sadeghian2021chronor} learns rotations parameterized by relation--time pairs. TeLM~\cite{xu2021temporal} uses multivector representations and an asymmetric geometric product. RotateQVS~\cite{chen2022rotateqvs} represents entities and relations as quaternions and models temporal changes in entities through quaternion rotations. LCGE~\cite{niu2023logic} incorporates temporal logic rules, and TeAST~\cite{li2023teast} maps relations onto an Archimedean spiral timeline.

Recent approaches further explore combinations of geometric transformations. HGE~\cite{pan2024hge} combines complex, dual, and split-complex subspaces with temporal attention. TCompoundE~\cite{ying2024tcompounde} combines time-specific and relation-specific geometric operations and analyzes their ability to capture relation patterns. TBicomR~\cite{nguyen2024tbicomr} and AttBiTi~\cite{le2025attbiti} use bicomplex embeddings for TKGC. BSCA instead uses the full non-commutative biquaternionic algebra and couples it with complex-valued attention to model changes in entity representations.

\section{Background}
\subsection{Problem Formulation}
A temporal knowledge graph $\mathcal{G}$ is a set of quadruples $(s, r, o, \tau)$, where $s, o \in \mathcal{E}$ are the subject (head) and object (tail), $r \in \mathcal{R}$ is a relation, and $\tau \in \mathcal{T}$ is a timestamp. The sets $\mathcal{E}$, $\mathcal{R}$, and $\mathcal{T}$ contain the entities, relation types, and timestamps, respectively. TKGE maps these elements to continuous representations and uses a time-aware scoring function $f(s, r, o, \tau)$ to assess the plausibility of each fact. Temporal knowledge graph completion (TKGC) answers queries such as $(s, r, ?, \tau)$ and $(?, r, o, \tau)$ by ranking candidate entities. A useful scoring function assigns higher scores to true facts than to incorrect candidates.

\subsection{Biquaternions}
Biquaternions were introduced to knowledge graph embedding by~\citet{guo2021bique}. A biquaternion belongs to a four-dimensional vector space over $\mathbb{C}$. We write a complex number as $c = c_r + c_i\mathbf{I}$, where $c_r, c_i \in \mathbb{R}$ and $\mathbf{I}^2 = -1$.

A biquaternion $q$ has the form:
\begin{equation}
	\begin{aligned}
		q = w + x\mathbf{i} + y\mathbf{j} + z\mathbf{k},
	\end{aligned}
\end{equation}
where $w, x, y, z \in \mathbb{C}$, and $\mathbf{i}, \mathbf{j}, \mathbf{k}$ are the imaginary units that satisfy the following multiplication rules:
\begin{equation}
	\begin{gathered}
		\mathbf{i j} = -\mathbf{j} \mathbf{i} = \mathbf{k},\quad \mathbf{j k} = -\mathbf{k j} = \mathbf{i},\quad \mathbf{k i} = -\mathbf{i k} = \mathbf{j}, \\
		\mathbf{i}^2 = \mathbf{j}^2 = \mathbf{k}^2 = -1.
	\end{gathered}
\end{equation}

Thus, the biquaternion $q$ can also be expressed as:
\begin{equation}
	\begin{aligned}
		q &= q_a + q_b \mathbf{I}, \\
		q_a &= w_a + x_a \mathbf{i} + y_a \mathbf{j} + z_a \mathbf{k}, \\
		q_b &= w_b + x_b \mathbf{i} + y_b \mathbf{j} + z_b \mathbf{k}.
	\end{aligned}
\end{equation}

The quaternion conjugate of $q$ is $\overline{q} = w - x\mathbf{i} - y\mathbf{j} - z\mathbf{k}$. Its complex conjugate is $q^* = w^* + x^*\mathbf{i} + y^*\mathbf{j} + z^*\mathbf{k}$, where $c^*$ denotes the complex conjugate of $c$.

\textbf{Basic Operations on Biquaternions.} Given two biquaternions $q_1 = w_1 + x_1 \mathbf{i} + y_1 \mathbf{j} + z_1 \mathbf{k}$ and $q_2 = w_2 + x_2 \mathbf{i} + y_2 \mathbf{j} + z_2 \mathbf{k}$, addition and subtraction are performed as follows:
\begin{equation}
	\begin{aligned}
		q_1 \pm q_2 = (w_1 \pm w_2) + (x_1 \pm x_2)\mathbf{i} + (y_1 \pm y_2)\mathbf{j} + (z_1 \pm z_2)\mathbf{k}.
	\end{aligned}
\end{equation}

\textbf{Hamilton Product.} The product of two biquaternions, $q_1 q_2$, is obtained via distributivity, respecting the multiplication properties of $\mathbf{i}$, $\mathbf{j}$, and $\mathbf{k}$:
\begin{equation}
	\begin{aligned}
		q_1 q_2 = & \quad w_1 w_2 - x_1 x_2 - y_1 y_2 - z_1 z_2 \\
		&+ (w_1 x_2 + x_1 w_2 + y_1 z_2 - z_1 y_2) \mathbf{i} \\
		&+ (w_1 y_2 - x_1 z_2 + y_1 w_2 + z_1 x_2) \mathbf{j} \\
		&+ (w_1 z_2 + x_1 y_2 - y_1 x_2 + z_1 w_2) \mathbf{k}.
	\end{aligned}
	\label{eq:hamilton}
\end{equation}
The Hamilton product is generally \textbf{non-commutative}, i.e., $q_1 q_2 \neq q_2 q_1$. It satisfies the conjugate reversal property: $\overline{q_1 q_2} = \overline{q_2}\,\overline{q_1}$. In the following, we use $\cdot$ to denote the Hamilton product where it aids readability.

\textbf{Norm of a Biquaternion.} Following~\cite{guo2021bique}, the (Hermitian) norm of a biquaternion $q$ is defined as:
\begin{equation}
	\|q\| = \sqrt{|w|^2 + |x|^2 + |y|^2 + |z|^2},
\end{equation}
where $|\cdot|$ denotes the modulus of a complex number. This yields a non-negative real value and reduces to the standard quaternion norm when $w, x, y, z \in \mathbb{R}$. A unit biquaternion is one where $\|q\| = 1$.

\subsection{Relation Patterns}
Following prior work~\cite{chen2022rotateqvs,sun2018rotate}, we consider the following relation patterns:\\
\textbf{Definition 1.} A relation $r$ is symmetric if $ \forall s, o, \tau, r(s, o,\tau) \wedge r(o, s, \tau)$ holds. \\
\textbf{Definition 2.} A relation $r$ is asymmetric if $ \forall s, o, \tau, r(s, o,\tau) \wedge \neg r(o, s, \tau)$ holds. \\
\textbf{Definition 3.} Relations $r_1$ and $r_2$ are inverse if $ \forall s, o, \tau, r_1(s, o,\tau) \wedge r_2(o, s, \tau)$ holds. \\
\textbf{Definition 4.} Relations $r_1$ and $r_2$ evolve from $\tau_1$ to $\tau_2$ if\\
$ \forall s, o, r_1(s, o, \tau_1) \wedge r_2(o, s, \tau_2)$ holds.

\section{Methodology}
\subsection{BSCA Model}
BSCA represents entities, relations, and timestamps in biquaternionic space. Motivated by the geometric transformations available in hypercomplex representations~\cite{zhang2019quaternion,chen2022rotateqvs}, we use biquaternions to model temporal facts. For a quadruple $(s, r, o, \tau)$, let $e_s$, $e_r$, $e_o$, and $e_\tau$ denote the subject, relation, object, and timestamp embeddings, respectively:
\begin{equation}
	\begin{aligned}
		e_s &= w_s + x_s\mathbf{i} + y_s\mathbf{j} + z_s\mathbf{k}, \\
		e_r &= w_r + x_r\mathbf{i} + y_r\mathbf{j} + z_r\mathbf{k}, \\
		e_o &= w_o + x_o\mathbf{i} + y_o\mathbf{j} + z_o\mathbf{k}, \\
		e_\tau &= w_\tau + x_\tau\mathbf{i} + y_\tau\mathbf{j} + z_\tau\mathbf{k},
	\end{aligned}
\end{equation}
where $e_s, e_r, e_o, e_\tau \in \mathbb{C}^{4 \times k}$. We parameterize each relation using a multiplicative component $e_r^\blacktriangle$ and an additive component $e_r^\blacktriangledown$. The multiplicative component encodes geometric transformations through the Hamilton product, whereas the additive component encodes translations in the embedding space. This separation allows the model to learn relation-specific transformations and semantic offsets independently. Similarly, each timestamp has three components: $e_\tau^\blacktriangle$ for multiplicative transformations, $e_\tau^\blacktriangledown$ for additive entity shifts, and $e_\tau^\blacklozenge$ for additive shifts in the joint relation--time representation.

We first incorporate temporal and relational information through translations. The vectors $e_\tau^\blacktriangledown$ and $e_r^\blacktriangledown$ shift the subject and object embeddings to form time-conditioned and relation-conditioned representations:
\begin{equation}
	\begin{aligned}
		e_{s\tau} = e_s + e_\tau^\blacktriangledown, \quad e_{sr} = e_s + e_r^\blacktriangledown, \\
		e_{o\tau} = e_o + e_\tau^\blacktriangledown, \quad e_{or} = e_o + e_r^\blacktriangledown.
	\end{aligned}
\end{equation}

We then condition the relation on time by defining $e_{r\tau} = e_r^\blacktriangle + e_\tau^\blacklozenge$. Applying the Hamilton product with the multiplicative timestamp component gives:
\begin{equation}
	\begin{aligned}
		e_{r\tau \tau} = e_{r\tau} \cdot e_\tau^\blacktriangle,
	\end{aligned}
\end{equation}
where $\cdot$ denotes the Hamilton product. The resulting representation $e_{r\tau \tau}$ supports circular rotations, hyperbolic rotations, and scaling transformations.

To make entity representations depend on their context, we use complex-valued attention to fuse $e_{s\tau}$ and $e_{sr}$:
\begin{equation}
	\begin{aligned}
		e_{s\tau r} &= a_\tau e_{s\tau} + a_r e_{sr}, \\
		(a_\tau, a_r) &= \text{softmax}(a^\top e_{s\tau}, a^\top e_{sr}),
	\end{aligned}
\end{equation}
where $a$ is a learnable complex-valued weight matrix. Complex-valued attention is motivated by the embedding space: complex weights can account for both magnitude and phase when combining temporal and relational features. The fused representation can therefore vary across timestamps and relations. We then apply the time-conditioned relation transformation to the subject: $e_{s\tau r}^{r\tau \tau} = e_{s\tau r} \cdot e_{r\tau \tau}$.

Following prior work~\cite{lacroix2018canonical,chen2022rotateqvs}, BSCA scores a fact using the inner product between the transformed subject and the object embedding:
\begin{equation}
	\begin{aligned}
		\phi(s,r,o,\tau) = \langle e_{s\tau r}^{r\tau\tau}, e_o \rangle.
	\end{aligned}
\end{equation}

For fixed numbers of entities, relations, and timestamps, the number of BSCA parameters grows linearly with the embedding dimension $k$, giving $O(k)$ parameter storage with respect to $k$.

\subsection{Optimization}
We train BSCA with a multiclass log-softmax objective and N3 regularization~\cite{lacroix2018canonical}. The regularizer is applied to the complex-valued embedding components to adapt it to the biquaternionic setting. We also use reciprocal learning: for each relation $r$, we introduce an inverse relation $r^{-1}$ so that both subject and object prediction contribute to training. The objective is:
\begin{equation}
	\begin{aligned}
		\mathcal{L}_\omega = & -\sum_{(s,r,o,\tau) \in \mathcal{G}} \left( \log p(o|s, r, \tau) + \log p(s|o, r^{-1}, \tau) \right) \\
		& + \lambda_{\mu} \sum_{i=1}^k \left(\|e_{s\tau r}\|_3^3 + \|e_{r\tau \tau}\|_3^3 + \|e_{o\tau r}\|_3^3\right),
	\end{aligned}
\end{equation}
where $p(o|s, r, \tau)$ is computed by applying softmax over all candidate objects, and $\lambda_{\mu}$ controls the strength of embedding regularization.

\subsection{Temporal Regularization}
To encourage smooth changes over time, we use the temporal regularization strategy of TComplEx~\cite{lacroix2019tensor}. This regularizer penalizes differences between embeddings at adjacent timestamps:
\begin{equation}
	\mathcal{L}_{\tau} = \sum_{i=1}^{N_{\tau}-1} \|e_{\tau_{i+1}} - e_{\tau_i}\|_p^p,
\end{equation}
where $N_{\tau}$ is the number of time steps. The penalty discourages abrupt changes between adjacent timestamp embeddings.

The total loss function of BSCA is defined as:
\begin{equation}
	\mathcal{L} = \mathcal{L}_{\omega} + \lambda_{\tau} \mathcal{L}_{\tau},
\end{equation}
where $\lambda_{\tau}$ balances the link prediction objective and the temporal smoothness penalty.

\subsection{Modeling Various Relation Patterns}
We consider symmetric, asymmetric, inverse, and temporally evolving relations. The following propositions state the corresponding representation capabilities; their proofs are provided in Appendix~\ref{app:proofs}. \\
\textbf{Proposition 1.} BSCA can model symmetric relations. See Appendix~\ref{app:prop1}. \\
\textbf{Proposition 2.} BSCA can model asymmetric relations. See Appendix~\ref{app:prop2}. \\
\textbf{Proposition 3.} BSCA can model inverse relations. See Appendix~\ref{app:prop3}. \\
\textbf{Proposition 4.} BSCA can model temporal evolution in relations. See Appendix~\ref{app:prop4}.

\begin{table*}[htb!]
	\centering
	\caption{Statistics of the experimental datasets.}
	\label{Table:datasets}
	\begin{tabular}{l|ccccc|ccc}
		\cmidrule[1pt](){1-9}
		\textbf{Datasets}  &\textbf{Entities}    &\textbf{Relations}    &\textbf{Timestamps}   &\textbf{Time Span}    &\textbf{Granularity}   &\textbf{Training} &\textbf{Validation}  &\textbf{Test}   \\
		\cmidrule{1-9}
		ICEWS14        &6,869  &230  &365  &A.D.2014  &1 day  &72,826 &8,941 &8,963 \\
		ICEWS05-15  &10,094  &251  &4,017  &A.D.2005-A.D.2015  &1 day  &368,962 &46,275 &46,092 \\
		GDELT    &500 &20 &366 &A.D.2015-A.D.2016 &1 day &2,735,685 &341,961 &341,961 \\
		YAGO11k &10,623 &10 &237 & 453 B.C.-A.D.2844 &1 year &16,406 &2,050 &2,051 \\
		Wikidata12k    &12,554  &24  &232 &A.D.1709-A.D.2018  &1 year  &539,286  &67,538 &63,110 \\
		\cmidrule[1pt](){1-9}
\end{tabular}
\end{table*}

\section{Experiments}
\subsection{Datasets}
We evaluate BSCA on five widely used TKG datasets: ICEWS14, ICEWS05-15~\cite{garcia2018learning}, GDELT, YAGO11k~\cite{garcia2018learning}, and Wikidata12k~\cite{dasgupta2018hyte}. Table~\ref{Table:datasets} summarizes their statistics. ICEWS14 and ICEWS05-15 are subsets of the Integrated Crisis Early Warning System (ICEWS) dataset~\cite{boschee2015icews}, which records temporal sociopolitical events. They cover 2014 and 2005--2015, respectively. Our GDELT dataset is a subset of the Global Database of Events, Language, and Tone~\cite{leetaru2013gdelt}, covering April 1, 2015, to March 31, 2016, with 500 entities and 20 relation types. For YAGO11k and Wikidata12k, we omit month and day information and merge adjacent years with few facts into intervals containing at least 300 facts. This preprocessing reduces temporal sparsity and variation in the number of facts per interval.

\subsection{Evaluation Metrics}
We evaluate link prediction using mean reciprocal rank (MRR) and Hits@$n$ for $n \in \{1,3,10\}$. MRR averages the reciprocal rank of the correct entity, while Hits@$n$ measures the proportion of queries for which the correct entity appears among the top $n$ candidates. For each test quadruple, we rank candidate subjects using $(s', r, o, \tau)$ and candidate objects using $(s, r, o', \tau)$. We apply temporal filtering when computing these ranks.

\subsection{Baselines}
We compare BSCA with the following TKGC baselines: TTransE~\cite{leblay2018deriving}, TA-DistMult~\cite{garcia2018learning}, HyTE~\cite{dasgupta2018hyte}, DE-SimplE~\cite{goel2020diachronic}, TComplEx and TNTComplEx~\cite{lacroix2019tensor}, ATiSE~\cite{xu2020temporal}, TeRo~\cite{xu2020tero}, ChronoR~\cite{sadeghian2021chronor}, TeLM~\cite{xu2021temporal}, BoxTE~\cite{messner2022temporal}, RotateQVS~\cite{chen2022rotateqvs}, TuckERTNT~\cite{shao2022tucker}, TLT-KGE($\mathbb{C}$) and TLT-KGE($\mathbb{Q}$)~\cite{zhang2022along}, LCGE~\cite{niu2023logic}, TeAST~\cite{li2023teast}, BDME~\cite{yue2023block}, SANe~\cite{li2022each}, and CEC-BD~\cite{yue2024complex}.

\begin{table*}[bt!]
	\caption{Link prediction results (\%) on five TKG datasets. Baseline results are taken from the original papers. \textbf{Bold} and \underline{underlining} indicate the best and second-best results, respectively. Dashes indicate unavailable results.}
	\label{Table:result_merged}
	\centering
	\resizebox{\textwidth}{!}{
		\begin{tabular}{ll|cccc|cccc|cccc|cccc|cccc}
			\cmidrule[1pt](){2-22}
			\multicolumn{2}{c}{\multirow{2}{*}{\vspace{-2.2mm}\hspace{-6mm}\textbf{Method}}}
			& \multicolumn{4}{c}{\textbf{ICEWS05-15}}
			& \multicolumn{4}{c}{\textbf{ICEWS14}}
			& \multicolumn{4}{c}{\textbf{GDELT}}
			& \multicolumn{4}{c}{\textbf{YAGO11k}}
			& \multicolumn{4}{c}{\textbf{Wikidata12k}} \\
			\cmidrule(lr){3-6} \cmidrule(lr){7-10} \cmidrule(lr){11-14} \cmidrule(lr){15-18} \cmidrule(lr){19-22}
			&   & \textbf{MRR} & \textbf{H@1} & \textbf{H@3} & \textbf{H@10}
			& \textbf{MRR} & \textbf{H@1} & \textbf{H@3} & \textbf{H@10}
			& \textbf{MRR} & \textbf{H@1} & \textbf{H@3} & \textbf{H@10}
			& \textbf{MRR} & \textbf{H@1} & \textbf{H@3} & \textbf{H@10}
			& \textbf{MRR} & \textbf{H@1} & \textbf{H@3} & \textbf{H@10} \\
			\cmidrule{2-22}
			& TTransE& 27.1 & 8.4 & -   & 61.6& 25.5 & 7.4 & -   & 60.1	& 11.5 & 0.0 & 16.0 & 31.8 & 10.8 & 2.0 & 15.0 & 25.1	& 17.2 & 9.6 & 18.4 & 32.9 \\
			& TA-DistMult & 47.4 & 34.6 & -   & 72.8 & 47.7 & 36.3 & -   & 68.6	& 20.6 & 12.4 & 21.9 & 36.5	& 15.5 & 9.8 & -   & 26.7	& 23.0 & 13.0 & -   & 46.1 \\
			& HyTE
			& 31.6 & 11.6 & 44.5 & 68.1
			& 29.7 & 10.8 & 41.6 & 65.5
			& 11.8 & 0.0 & 16.5 & 32.6
			& 13.6 & 3.3 & -    & 29.8
			& 18.0 & 9.8 & 19.7 & 33.3 \\
			& DE-SimplE
			& 51.3 & 39.2 & 57.8 & 74.8
			& 52.6 & 41.8 & 59.2 & 72.5
			& 23.0 & 14.1 & 24.8 & 40.3
			& - & - & - & -
			& - & - & - & - \\
			& TComplEx
			& 66.0 & 59.0 & 71.0 & 80.0
			& 61.0 & 53.0 & 66.0 & 76.0
			& 34.0 & 24.9 & 36.1 & 49.8
			& 18.5 & 12.7 & 18.3 & 30.7
			& 33.1 & 23.3 & 35.7 & 53.9 \\
			& TNTComplEx
			& 67.0 & 59.0 & 71.0 & 81.0
			& 62.0 & 52.0 & 66.0 & 76.0
			& 34.9 & 25.8 & 37.3 & 50.2
			& 18.0 & 11.0 & - & 31.2
			& 30.1 & 19.7 & - & 50.6 \\
			& ATiSE
			& 51.9 & 37.8 & 60.6 & 79.4
			& 55.0 & 43.6 & 62.9 & 75.0
			& - & - & - & -
			& 18.5 & 12.6 & 18.9 & 30.1
			& 25.2 & 14.8 & 28.8 & 46.2 \\
			& TeRo
			& 58.6 & 46.9 & 66.8 & 79.5
			& 56.2 & 46.8 & 62.1 & 73.2
			& 24.5 & 15.3 & 26.4 & 42.0
			& 18.7 & 12.1 & 19.7 & 31.9
			& 29.9 & 19.8 & 32.9 & 50.7 \\
			& ChronoR
			& 67.5 & 59.6 & 72.3 & 82.0
			& 62.5 & 54.7 & 66.9 & 77.3
			& - & - & - & -
			& - & - & - & -
			& - & - & - & - \\
			& TeLM
			& 67.8 & 59.9 & 72.8 & 82.3
			& 62.5 & 54.5 & 67.3 & 77.4
			& 35.0 & 26.1 & 37.5 & 50.4
			& 19.1 & 12.9 & 19.4 & 32.1
			& 33.2 & 23.1 & 36.0 & 54.2 \\
			& BoxTE
			& 66.7 & 58.2 & 71.9 & 82.0
			& 61.3 & 52.8 & 66.4 & 76.3
			& 35.3 & 26.9 & 37.7 & 51.1
			& - & - & - & -
			& - & - & - & - \\
			& RotateQVS
			& 63.3 & 52.9 & 70.9 & 81.3
			& 59.1 & 50.7 & 64.2 & 75.4
			& 27.0 & 17.5 & 29.3 & 45.8
			& - & - & - & -
			& - & - & - & - \\
			& TuckERTNT
			& 63.8 & 55.9 & 68.6 & 78.3
			& 60.4 & 52.1 & 65.5 & 75.3
			& \underline{38.1} & \underline{28.3} & \underline{40.1} & \underline{54.4}
			& - & - & - & -
			& - & - & - & - \\
			& TLT-KGE($\mathbb{C}$)
			& 68.6 & 60.7 & 73.5 & 83.1
			& 63.0 & 54.9 & 67.8 & 77.7
			& 35.6 & 26.7 & 38.5 & 53.2
			& - & - & - & -
			& - & - & - & - \\
			& TLT-KGE($\mathbb{Q}$)
			& \underline{69.0} & \underline{60.9} & \underline{74.1} & \underline{83.5}
			& 63.4 & 55.1 & 68.4 & \underline{78.6}
			& 35.8 & 26.5 & 38.8 & 54.3
			& - & - & - & -
			& - & - & - & - \\
			& LCGE
			& 61.8 & 51.4 & 68.1 & 81.2
			& 61.6 & 53.2 & 66.7 & 77.5
			& - & - & - & -
			& - & - & - & -
			& - & - & - & - \\
			& TeAST
			& 68.3 & 60.4 & 73.2 & 82.9
			& 63.7 & \underline{56.0} & 68.2 & 78.2
			& 37.1 & 28.3 & 40.1 & 54.4
			& - & - & - & -
			& - & - & - & - \\
			& BDME
			& - & - & - & -
			& 63.5 & 55.5 & 68.3 & 77.8
			& 27.8 & 19.1 & 29.9 & 44.8
			& - & - & - & -
			& - & - & - & - \\
			& SANe
			& 68.3 & 60.5 & 73.4 & 82.3
			& \underline{63.8} & 55.8 & \underline{68.8} & 78.2
			& - & - & - & -
			& - & - & - & -
			& - & - & - & - \\
			& CEC-BD
			& 68.1 & 60.2 & 73.0 & 82.5
			& 63.3 & 55.4 & 68.0 & 77.7
			& 29.6 & 20.1 & 33.4 & 46.5
			& \textbf{21.2} & \textbf{15.4} & \underline{21.5} & \underline{33.9}
			& \underline{33.9} & \underline{24.1} & \underline{36.9} & \underline{54.3} \\
			\cmidrule[0.6pt](){2-22}
			& \textbf{BSCA}
			& \textbf{69.3} & \textbf{61.3} & \textbf{74.3} & \textbf{83.8}
			& \textbf{64.7} & \textbf{56.6} & \textbf{69.4} & \textbf{79.6}
			& \textbf{52.1} & \textbf{44.6} & \textbf{56.1} & \textbf{65.4}
			& \underline{20.9} & \underline{14.2} & \textbf{21.7} & \textbf{34.4}
			& \textbf{34.3} & \textbf{24.3} & \textbf{37.2} & \textbf{55.6} \\
			\cmidrule[1pt](){2-22}
		\end{tabular}
	}
\end{table*}

\begin{table*}[htbp]
	\caption{Ablation results (\%) on four TKG datasets. The full BSCA model is shown in bold for reference; w/o denotes removal of the indicated component.}
	\label{Table:ablation}
	\centering
	\resizebox{\textwidth}{!}{
		\begin{tabular}{ll|cccc|cccc|cccc|cccc}
			\cmidrule[1pt](){2-18}
			\multicolumn{2}{c}{\multirow{2}{*}{\vspace{-2.2mm}\hspace{-6mm}\textbf{Method}}} & \multicolumn{4}{c}{\textbf{ICEWS05-15}} & \multicolumn{4}{c}{\textbf{ICEWS14}} & \multicolumn{4}{c}{\textbf{YAGO11k}}& \multicolumn{4}{c}{\textbf{Wikidata12k}} \\
			\cmidrule(lr){3-6} \cmidrule(lr){7-10} \cmidrule(lr){11-14} \cmidrule(lr){15-18}
			&                   &\textbf{MRR} &\textbf{Hits@1}   &\textbf{Hits@3}    &\textbf{Hits@10}  &\textbf{MRR} &\textbf{Hits@1}   &\textbf{Hits@3}    &\textbf{Hits@10} &\textbf{MRR} &\textbf{Hits@1}   &\textbf{Hits@3}    &\textbf{Hits@10} &\textbf{MRR} &\textbf{Hits@1}   &\textbf{Hits@3}    &\textbf{Hits@10} \\
			\cmidrule{2-18}
			&\textbf{BSCA} &\textbf{69.3} &\textbf{61.3} &\textbf{74.3} &\textbf{83.8} &\textbf{64.7} &\textbf{56.6} &\textbf{69.4} &\textbf{79.6} &\textbf{20.9} &\textbf{14.2} &\textbf{21.7} &\textbf{34.4} &\textbf{34.3} &\textbf{24.3} &\textbf{37.2} &\textbf{55.6}\\
			&w/o $e_{\tau}^\blacktriangledown$ &68.7 &60.7 &73.7 &83.1 &64.3 &56.2 &69.2 &79.4 &20.4 &13.7 &21.0 &33.3 &33.6 &23.3 &36.9 &55.4 \\
			&w/o $e_r^\blacktriangledown$ & 68.4 &60.3 &73.4 &83.2 &64.2 &55.9 &69.2 &79.6 &20.3 &13.7 &21.0 &34.5 &33.7 &23.2 &37.3 &55.7 \\
			&w/o $e_{\tau}^\blacktriangledown + e_r^\blacktriangledown$ &68.9 &60.8 &74.0 &83.8 &63.9 &55.7 &68.9 &79.4 &20.1 &13.6 &20.6 &33.3 &33.7 &23.0 &37.2 &56.1\\
			&w/o $e_\tau^{\blacktriangle}$ & 68.8 & 60.8 &73.8 & 83.3 &64.3 &56.2 &68.9 &79.5 &20.5 &13.7 &21.2 &34.8 &34.2 &23.5 &37.6 &56.2 \\
			&w/o Attention & 68.5 &60.6 &73.6 &83.0 &64.2 &55.9 &69.2 &79.6 &20.3 &13.5 &21.1 &34.4 &33.6 &23.2 &36.8 &55.6 \\
			&w/o Biquaternionic &68.6 &60.6 &73.6 &83.0 &64.3 &56.2 &68.9 &79.1 &19.8 &13.7 &19.8 &32.9 &33.5 &23.5 &36.3 &54.6 \\
			\cmidrule[1pt](){2-18}
		\end{tabular}
	}
\end{table*}

\subsection{Experimental Setup}
We implement BSCA in PyTorch and run all experiments on an NVIDIA GeForce RTX 3090 GPU with 24\,GB of memory. We use Adagrad with a learning rate of 0.1 and a batch size of 6000, and select hyperparameters by grid search on the validation set. The main experiments use an embedding dimension of 2000 and run for up to 150 epochs. The separate hyperparameter study explores dimensions from 500 to 8000, and the convergence study extends training to 300 epochs. We check stability across multiple random seeds and report results from a single representative run. The selected regularization weights are: ICEWS14 ($\lambda_\mu=0.008, \lambda_\tau=0.01$), ICEWS05-15 ($\lambda_\mu=0.002, \lambda_\tau=0.05$), GDELT ($\lambda_\mu=5e-5, \lambda_\tau=0.2$), YAGO11k ($\lambda_\mu=0.1, \lambda_\tau=0.009$), and Wikidata12k ($\lambda_\mu=0.1, \lambda_\tau=5e-4$).

\begin{figure*}[htbp]
	\centering
	\begin{minipage}{0.3\linewidth}
		\centerline{\includegraphics[width=\textwidth]{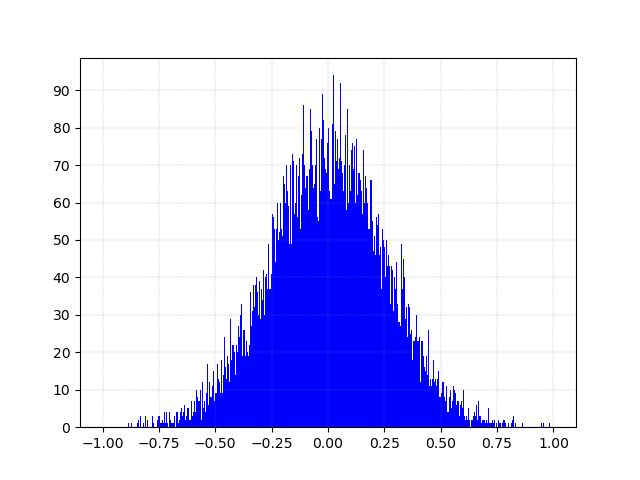}}
		\centerline{(a)~$e_{r\tau}$}
	\end{minipage}
	\begin{minipage}{0.3\linewidth}
		\centerline{\includegraphics[width=\textwidth]{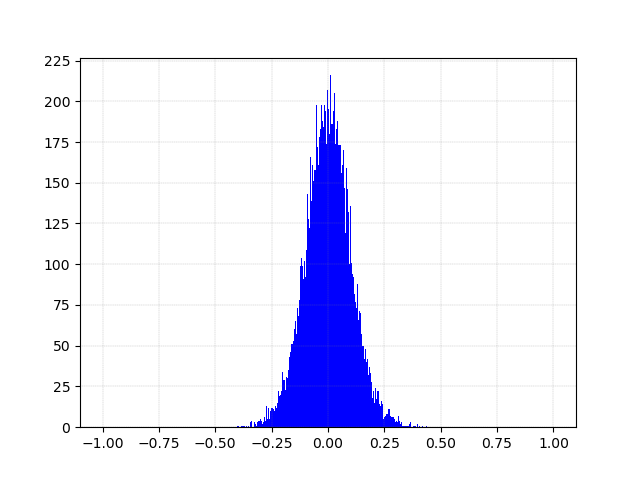}}
		\centerline{(b)~$e_{\tau}^\blacktriangle$}
	\end{minipage}
	\begin{minipage}{0.3\linewidth}
		\centerline{\includegraphics[width=\textwidth]{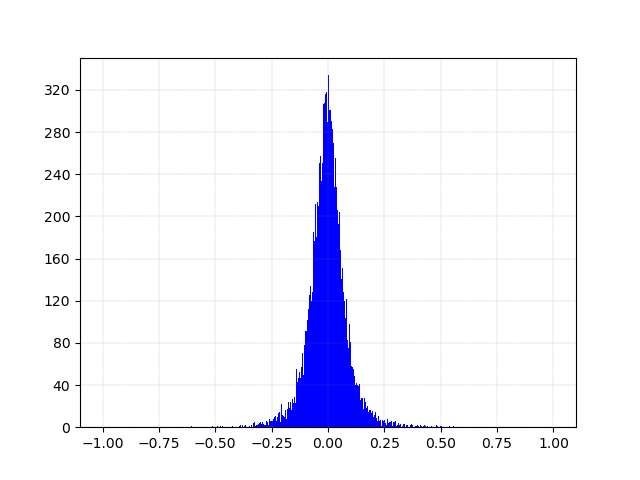}}
		\centerline{(c)~$e_{r\tau \tau}$}
	\end{minipage}
	\caption{Histograms of learned relation and timestamp representations for the relation ``Consult'' at timestamp $\tau=$ 2014-03-29.}
	\label{fig:visu}
\Description{Three histograms show the joint relation--time representation, the multiplicative timestamp component, and their Hamilton product for Consult on March 29, 2014.}
\end{figure*}

\begin{figure}[htbp]
	\centering
	\begin{minipage}{0.49\linewidth}
		\centerline{\includegraphics[width=\textwidth]{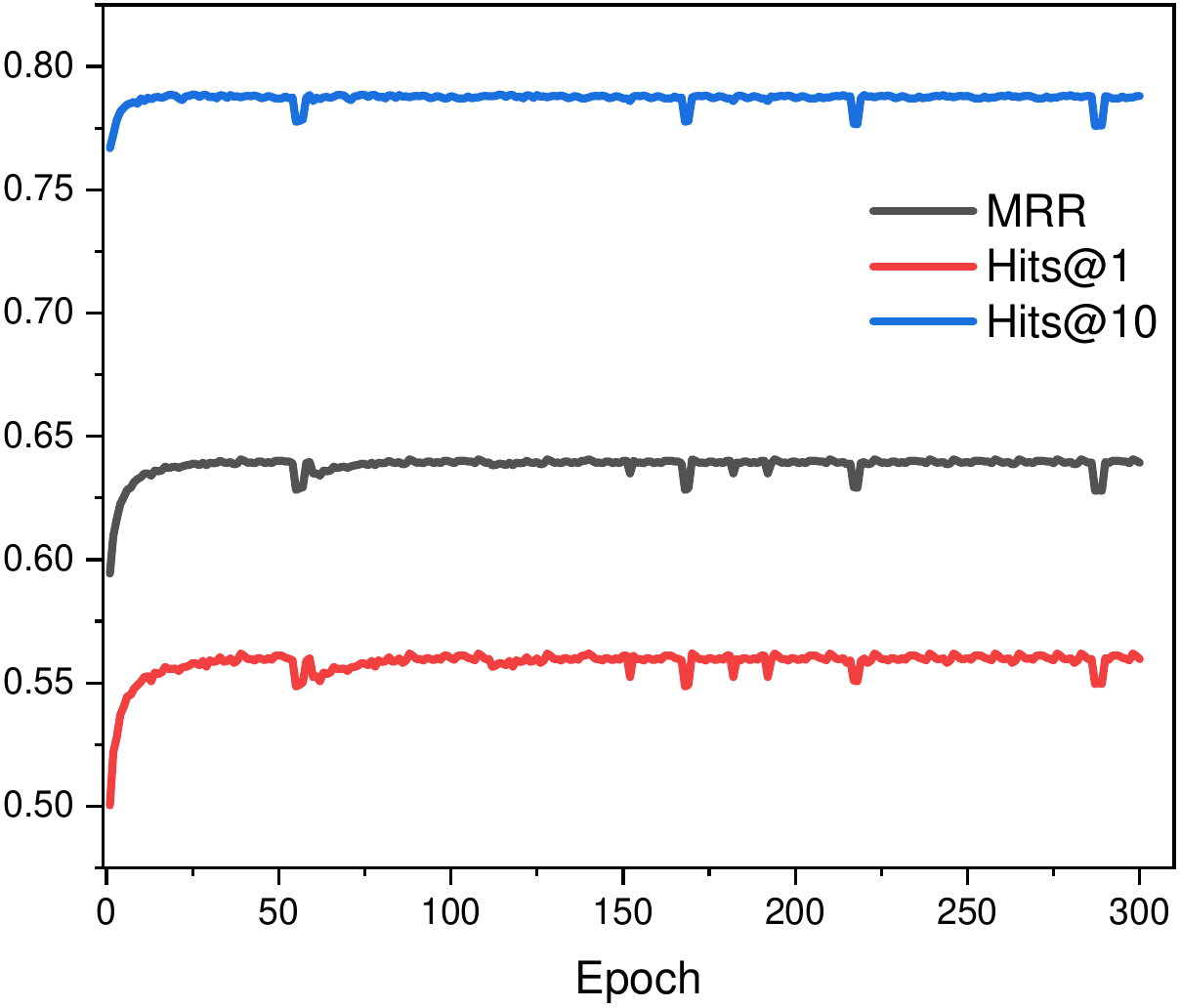}}
	\end{minipage}
	\begin{minipage}{0.49\linewidth}
		\centerline{\includegraphics[width=\textwidth]{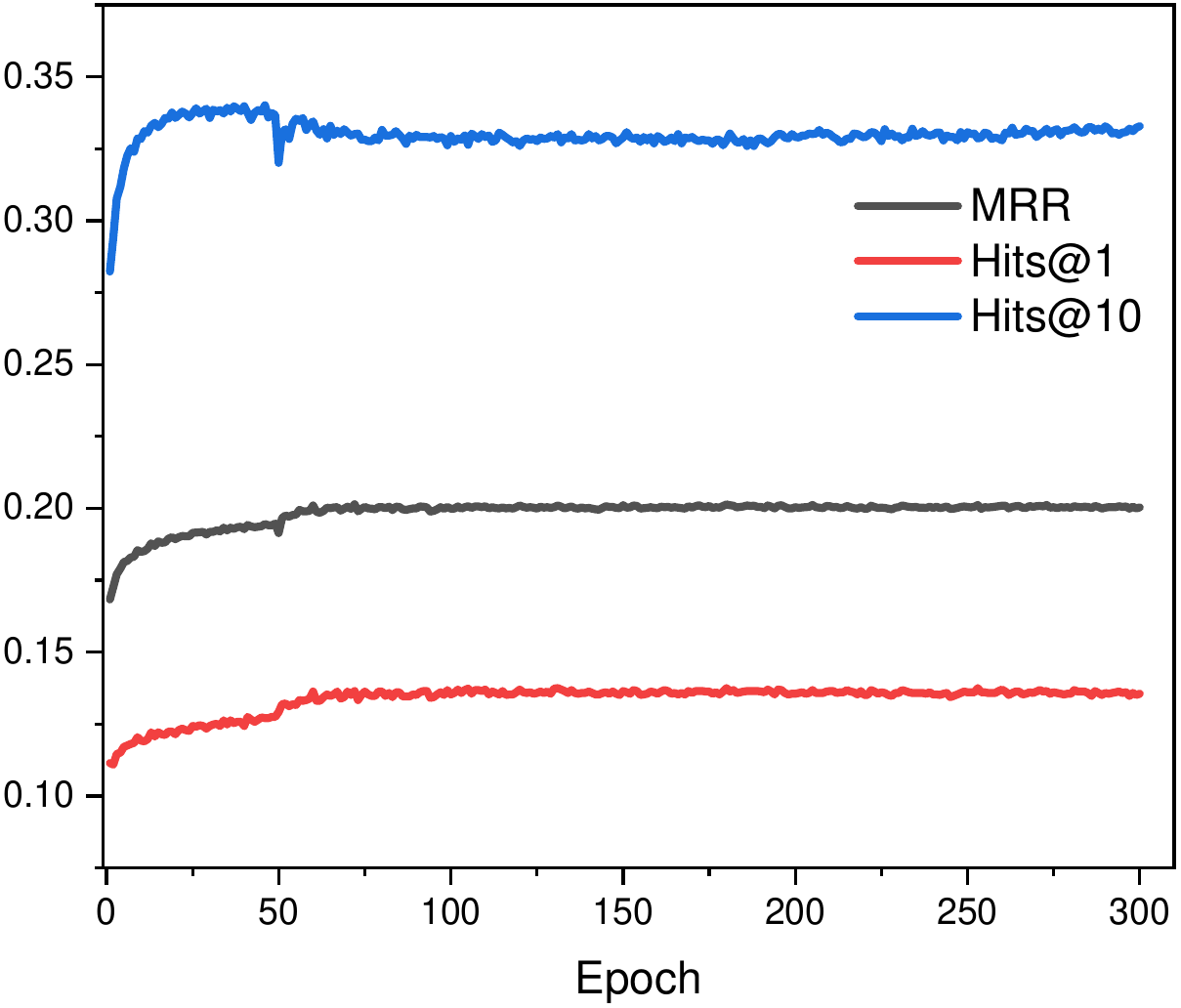}}
	\end{minipage}
	\caption{Training curves over 300 epochs on ICEWS14 and YAGO11k.}
	\label{fig:training_curve}
\Description{Two panels show BSCA link prediction performance over 300 training epochs on ICEWS14 and YAGO11k.}
\end{figure}

\subsection{Main Results}
Table~\ref{Table:result_merged} reports link prediction results on all five datasets. BSCA achieves the highest reported scores on all four metrics for ICEWS05-15, ICEWS14, GDELT, and Wikidata12k. On YAGO11k, it achieves the best Hits@3 and Hits@10, while CEC-BD has higher MRR and Hits@1. These results support the effectiveness of combining biquaternionic transformations with context-dependent entity representations.

The largest gain occurs on GDELT, where BSCA achieves an MRR of 52.1\%, compared with 38.1\% for TuckERTNT, the strongest baseline by MRR in this comparison. This is a gain of 14.0 percentage points, or approximately 37\% relative improvement. GDELT contains approximately 2.7 million training facts but only 500 entities and 20 relation types. This density may provide more observations from which to learn relation- and time-dependent transformations. We interpret this as a possible explanation for the larger gain, rather than direct evidence that density causes the improvement. The reported evaluation uses temporal filtering. Results on YAGO11k and Wikidata12k also show that BSCA remains competitive when timestamps have coarser granularity and span longer periods.

\begin{figure*}[!t]
	\centering
	\includegraphics[width=\linewidth]{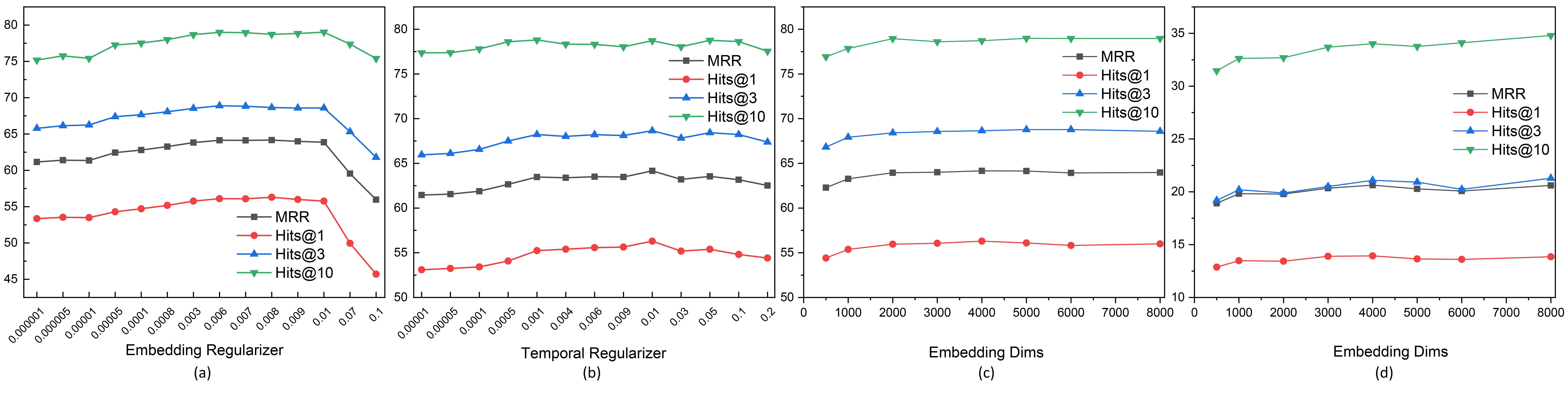}
	\caption{Sensitivity to embedding regularization (a) and temporal regularization (b) on ICEWS14, and to embedding dimension on ICEWS14 (c) and YAGO11k (d).}
	\label{fig:combined}
\Description{Four panels compare link prediction performance across embedding regularization weights, temporal regularization weights, and embedding dimensions on ICEWS14 and YAGO11k.}
\end{figure*}

\subsection{Ablation Studies}
We evaluate the contributions of the additive components, the multiplicative timestamp component, attention, and the biquaternionic representation. Table~\ref{Table:ablation} reports results on ICEWS05-15, ICEWS14, YAGO11k, and Wikidata12k.

Removing either $e_r^\blacktriangledown$ or $e_{\tau}^\blacktriangledown$ reduces MRR on all four datasets, supporting the contribution of relation-specific and time-specific translations. Removing attention also lowers MRR across the four datasets. These reductions are modest, and some ablated variants match or exceed the full model on individual Hits metrics. The results therefore support a consistent MRR benefit, rather than an improvement in every metric.

Several ablated variants also remain competitive with or outperform TComplEx and TNTComplEx on the reported datasets, suggesting that the model's performance is not attributable to a single component.

\subsection{Analysis of Relation Embeddings}
To examine the learned relation transformations, we visualize representations associated with the symmetric fact \textit{(Afghanistan, Consult, Iran, 2014-03-29)}. Figure~\ref{fig:visu} shows histograms of the corresponding relation and timestamp embeddings.

Figure~\ref{fig:visu}(c) shows that the components of $e_{r\tau\tau}$ are concentrated near zero for this example. This observation describes the learned transformation for one symmetric relation at one timestamp. It does not, by itself, establish a general condition for symmetry or show that temporal information has little effect on the relation. The visualization should therefore be interpreted as a qualitative example alongside the formal analysis.

\subsection{Hyperparameter Analysis}
We examine sensitivity to regularization weights and embedding dimension in Figure~\ref{fig:combined}.

On ICEWS14, the best embedding and temporal regularization weights are approximately $\lambda_\mu=0.008$ and $\lambda_\tau=0.01$, respectively, as shown in Figure~\ref{fig:combined}(a) and (b). Performance varies more strongly with the temporal regularization weight over the tested ranges, indicating that this hyperparameter warrants careful tuning.

We vary the embedding dimension from 500 to 8000 on ICEWS14 and YAGO11k. As shown in Figure~\ref{fig:combined}(c) and (d), performance improves up to a dimension of 4000, with smaller gains beyond 2000. We use 2000 in the main experiments to balance predictive performance and computational cost.

The 300-epoch runs in Figure~\ref{fig:training_curve} examine convergence beyond the training budget used for the main experiments. Performance generally stabilizes after approximately 70 epochs on both datasets. ICEWS14 exhibits fluctuations around epochs 60 and 150, followed by a return to a similar performance level.

\begin{table}[htbp]
	\centering
	\caption{MRR (\%) with and without attention for selected ICEWS14 relations and entities. Bold indicates the higher score in each row.}
	\label{Table:attention}
	\resizebox{0.95\linewidth}{!}{
	\begin{tabular}{llcc}
		\toprule
		\textbf{Relation} & \textbf{Pattern} & \textbf{w/o Attn} & \textbf{w/ Attn} \\
		\midrule
		Consult & Symmetric & 87.07 & \textbf{87.11} \\
		Discuss by telephone & Symmetric & \textbf{87.53} & 87.21 \\
		Engage in negotiation & Symmetric & 89.97 & \textbf{90.68} \\
		\midrule
		Make a visit & Asymmetric & 89.47 & \textbf{89.90} \\
		Make optimistic comment & Asymmetric & 52.82 & \textbf{55.09} \\
		Reject & Asymmetric & \textbf{57.92} & 57.90 \\
		\bottomrule
		\midrule
		\textbf{Entity} & \textbf{Time Span (Days)} & \textbf{w/o Attn} & \textbf{w/ Attn} \\
		\midrule
		China & 364 & 76.75 & \textbf{76.80} \\
		Japan & 356 & 81.80 & \textbf{83.65} \\
		Malaysia & 344 & 80.96 & \textbf{82.33} \\
		Romania & 356 & 76.51 & \textbf{79.71} \\
		Foreign Affairs (US) & 341 & 52.34 & \textbf{54.00} \\
		\bottomrule
	\end{tabular}}
\end{table}

\subsection{Attention Analysis}
Attention adaptively weights time-conditioned and relation-conditioned entity representations. Table~\ref{Table:attention} shows that it improves MRR for four of the six selected relations, including ``Engage in negotiation'' and ``Make a visit''. The effect is slightly negative for ``Reject'' and ``Discuss by telephone'', indicating that the benefit varies by relation. This variation is consistent with the modest aggregate gains in Table~\ref{Table:ablation}.

Attention improves MRR for all five selected entities, each of which appears over an extended period. These results support the usefulness of context-dependent fusion for these entities, although they do not isolate the mechanism responsible for the improvement.

\begin{table}[htbp]
	\centering
	\caption{Comparison of model complexity and training efficiency. All experiments were conducted on the ICEWS14 dataset using a single NVIDIA GeForce RTX 3090 (24GB).}
	\label{table:limitation}
	\begin{tabular}{lcc}
		\toprule
		\textbf{Method} & \textbf{\# Params (M)} & \textbf{Train Time} \\
		\midrule
		TComplEx & 31.81 & 7.9 min \\
		TNTComplEx & 32.65 & 8.4 min \\
		TeAST & 33.28 & 8.2 min \\
		TLT-KGE($\mathbb{C}$) & 58.56 & 8.2 min \\
		TLT-KGE($\mathbb{Q}$) & 31.28 & 14.1 min \\
		TeLM & 63.63 & 17.1 min \\
		\midrule
		\textbf{BSCA (Ours)} & \textbf{38.41} & \textbf{9.5 min} \\
		\bottomrule
	\end{tabular}
\end{table}

\section{Model Efficiency Analysis}
Table~\ref{table:limitation} compares parameter counts and training times on ICEWS14. BSCA uses 38.41 million parameters and takes 9.5 minutes to train in the reported setting. Its biquaternionic transformations and attention introduce additional cost relative to TComplEx and TNTComplEx.

Compared with TeLM, BSCA uses approximately 40\% fewer parameters and reduces training time from 17.1 to 9.5 minutes. BSCA also uses fewer parameters than TLT-KGE($\mathbb{C}$), although its training time is longer (9.5 versus 8.2 minutes). These comparisons indicate a useful balance between predictive performance and computational cost in the tested setting.

\section{Conclusion}
We presented BSCA, a TKGE model that combines circular and hyperbolic transformations in biquaternionic space with complex-valued attention for context-dependent entity representations. Across five benchmarks, BSCA achieves strong link prediction performance, with the largest MRR gain on GDELT. Ablation studies show consistent MRR benefits from the additive components and attention, while the efficiency comparison indicates moderate computational cost. These findings support further investigation of hypercomplex representations for temporal knowledge graph completion.

%% Acknowledgments, if applicable
%\begin{acks}
%...
%\end{acks}

%% Bibliography
\bibliographystyle{unsrtnat}
\bibliography{uai2026-template}

%% Language assistance disclosure retained from the source manuscript
%\section*{Generative AI Usage Disclosure}
%The authors used large language model tools to assist with grammar checking and language polishing of the manuscript. All scientific content, experimental design, quantitative results, mathematical derivations, and technical contributions are the responsibility of the authors.

\clearpage
\appendix
\section{Proofs of Propositions}
\label{app:proofs}

Throughout this appendix, $\cdot$ denotes the Hamilton product and $\circ$ denotes the inner product of biquaternion vectors.

\subsection{Proof of Proposition 1}
\label{app:prop1}
Expanding the scoring function gives:
\begin{equation}
	\begin{aligned}
		\phi(s, r, o, \tau) = \langle e_{s\tau r}^{r\tau\tau}, e_o \rangle = \langle e_{s\tau r} \cdot e_{r\tau \tau}, e_o \rangle = e_{s\tau r} \cdot e_{r\tau \tau} \circ e_o.
	\end{aligned}
\end{equation}
For a symmetric relation, Definition~1 gives $r(s, o, \tau) \wedge r(o, s, \tau)$. The corresponding score equality is:
\begin{equation}
	\phi(s, r, o, \tau) = \phi(o, r, s, \tau).
\end{equation}
Substituting and factoring out the common term $e_{r\tau \tau}$:
\begin{equation}
	\begin{aligned}
		\phi(s, r, o, \tau) = \phi(o, r, s, \tau) &\Leftrightarrow \\
		e_{s\tau r} \cdot e_{r\tau \tau} \circ e_o = e_{o\tau r} \cdot e_{r\tau \tau} \circ e_s &\Leftrightarrow \\
		e_s \circ e_o = e_{r\tau \tau}^{-1} e_{r\tau \tau} \, e_o \circ e_s.
	\end{aligned}
\end{equation}
When $e_{r\tau \tau}$ is invertible, the condition $e_s \circ e_o = e_o \circ e_s$ holds, which is satisfied if $e_s \circ e_o$ is symmetric. Therefore, BSCA can model symmetric patterns when the biquaternion $e_{r\tau \tau}$ is invertible.

\subsection{Proof of Proposition 2}
\label{app:prop2}
For an asymmetric relation, Definition~2 gives:
\begin{equation}
	r(s, o, \tau) \land \neg r(o, s, \tau).
\end{equation}
This implies $\phi(s, r, o, \tau) \neq \phi(o, r, s, \tau)$. Expanding the scoring function gives:
\begin{equation}
	e_{s\tau r} \cdot e_{r\tau\tau} \circ e_o \neq e_{o\tau r} \cdot e_{r\tau\tau} \circ e_s.
\end{equation}
Suppose $e_{r\tau\tau}$ were invertible. Right-multiplying by $e_{r\tau\tau}^{-1}$ (on the inner product terms) would yield $e_{s\tau r} \circ e_o = e_{o\tau r} \circ e_s$, making the score function symmetric---directly contradicting the asymmetry requirement. Therefore, $e_{r\tau\tau}$ must be non-invertible for asymmetric relations, and BSCA can represent such patterns under this condition.

\subsection{Proof of Proposition 3}
\label{app:prop3}
For inverse relations, Definition~3 gives $r_1(s, o, \tau) \land r_2(o, s, \tau)$. The corresponding score equality is:
\begin{equation}
	\begin{aligned}
		\phi(s, r_1, o, \tau) &= \phi(o, r_2, s, \tau) \Leftrightarrow \\
		e_s \circ e_o &= e_{r_1\tau \tau}^{-1} \, e_{r_2\tau \tau} \, e_o \circ e_s \Leftrightarrow \\
		e_{r_1\tau \tau}^{-1} \, e_{r_2\tau \tau} &= \mathbf{I} \quad \text{or} \quad e_{r_2\tau \tau}^{-1} \, e_{r_1\tau \tau} = \mathbf{I}.
	\end{aligned}
\end{equation}
Therefore, BSCA can model inverse patterns when $e_{r_1\tau \tau}$ and $e_{r_2\tau \tau}$ are inverses of each other.

\subsection{Proof of Proposition 4}
\label{app:prop4}
For temporal evolution, Definition~4 gives $r_1(s, o, \tau_1) \land r_2(o, s, \tau_2)$. The corresponding score equality is:
\begin{equation}
	\begin{aligned}
		\phi(s, r_1, o, \tau_1) &= \phi(o, r_2, s, \tau_2) \Leftrightarrow \\
		e_{r_1\tau_1 \tau_1} e_s \circ e_o &= e_{r_2\tau_2 \tau_2} e_o \circ e_s.
	\end{aligned}
\end{equation}
The derivation continues as follows:
\begin{equation}
	\begin{aligned}
		e_s \circ e_o &= e_{r_1\tau_1 \tau_1}^{-1} e_{r_2\tau_2 \tau_2} e_o \circ e_s \Rightarrow \\
		e_{r_1\tau_1 \tau_1}^{-1} e_{r_2\tau_2 \tau_2} &= \mathbf{I} \quad \text{or} \quad e_{r_2\tau_2 \tau_2}^{-1} e_{r_1\tau_1 \tau_1} = \mathbf{I}.
	\end{aligned}
\end{equation}
Therefore, BSCA can model temporal evolution patterns when $e_{r_1\tau_1 \tau_1}$ and $e_{r_2\tau_2 \tau_2}$ are inverses of each other.

\end{document}